\documentclass{article}
\usepackage[preprint]{neurips_2019}

\usepackage[utf8]{inputenc} 
\usepackage[T1]{fontenc}    
\usepackage{url}  
\usepackage{booktabs}       
\usepackage{adjustbox}
\usepackage{amsfonts}       
\usepackage{nicefrac}       
\usepackage{microtype}      
\usepackage{natbib}
\usepackage{wrapfig}
\usepackage{graphicx}
\usepackage{amsmath}
\usepackage{booktabs}
\usepackage{algpseudocode}
\usepackage{algorithm}
\usepackage{tikz}

\usepackage{chngcntr}
\usepackage{makecell}
\usepackage{caption}
\usepackage{subcaption}

\usepackage{bm}
\usepackage{amssymb}

\floatname{algorithm}{Algorithm}

\newtheorem{definition}{Definition}

\newcommand{\argmin}{\operatornamewithlimits{arg\min}}
\usepackage[colorlinks=true,linkcolor=red,citecolor=blue]{hyperref}

\usepackage{xcolor}

\usepackage{enumitem}
\setlist{leftmargin=8.5mm}
\setlist[itemize]{leftmargin=5.0mm}

\title{Generative Parameter Sampler
For Scalable Uncertainty Quantification}

\author{%
 Minsuk Shin\\
  Harvard Data Science Initiative and Department of Statistics\\
  Harvard University\\
  Cambridge, MA 02138 \\
  \texttt{minsuk000@gmail.com} \\
   \And
   Young Lee\\
  Department of Statistics\\
  Harvard University\\
  Cambridge, MA 02138 \\
  \texttt{younglee@fas.harvard.edu} \\
     \And
   Jun S. Liu\\
  Department of Statistics\\
  Harvard University\\
  Cambridge, MA 02138 \\
  \texttt{jliu@stat.harvard.edu} \\
}

\begin{document}
\maketitle
\begin{abstract}
    Uncertainty quantification has been a core of the statistical machine learning, but its computational bottleneck has been a serious challenge  for both Bayesians and frequentists. We propose a model-based framework in quantifying uncertainty, called \emph{predictive-matching Generative Parameter Sampler} (GPS). This procedure considers an \emph{Uncertainty Quantification} (UQ) distribution, on the targeted parameter, which matches the corresponding predictive distribution to the observed data. This framework adopts a hierarchical modeling perspective such that each observation is modeled by an individual parameter. This individual parameterization permits the resulting inference to be computationally scalable and robust to outliers. Our approach is illustrated for linear models, Poisson processes, and deep neural networks for classification. The results show that the GPS is successful in providing uncertainty quantification as well as additional flexibility beyond what is allowed by classical statistical procedures under the postulated statistical models.
\end{abstract}
\section{Introduction}
\label{introduction}






There have been significant efforts focused on identifying and quantifying  uncertainties  associated  with  machine learning and data
science procedures 
\citep{gal2016dropout,kendall2017uncertainties,srivastava2015wasp}. 
However, recent progresses in machine learning models such as deep neural networks are unsatisfactory at quantifying uncertainty and tend to produce overconfident predictions \citep{balaji:2017}. 
This article proposes a model-based framework for quantifying uncertainty in the inference of parameters. Uncertainty quantification (UQ) aims to provide a paradigm where uncertainty can be reviewed and ideally, assessed, in a manner relevant to researchers using the predictive models. This is conducted by searching an \emph{Uncertainty Quantification} (UQ)  distribution of parameters in interest that minimizes a \emph{ distance} between the predictive distribution based on the UQ distribution and the empirical distribution of the observed data.  Since directly evaluating the UQ distribution (e.g., the density function) is computationally challenging, we consider a parameter generator that generates parameter samples that follow the minimizer UQ distribution to ease the computational difficulty. In this sense, we call the proposed procedure  \emph{predictive-matching Generative Parameter Sampler} (GPS), which is implemented by a stochastic optimization algorithm so that the computation is scalable for large-sized data sets. Our main contributions are summarized as follows:

\begin{enumerate}
    \item[\textbf{1.}] We propose a new framework to quantify uncertainty of models and parameters according to 
    predictive-matching with the observed data.
    The UQ based on the GPS is predictive optimal in a sense that the resulting predictive distribution \eqref{eq:pred} based on the UQ distribution (defined in \eqref{eq:GPS}) is as close as possible to the empirical distribution of the observed data. 

        
    \item[\textbf{2.}]  We show that GPS induces a set of \emph{fully} independent structure based on the individual parameterization (Figure \ref{fig:gps-vs-standard}\,(a)) in contrast to the global parameterization for standard Bayesian (or frequentist) models (Figure \ref{fig:gps-vs-standard}\,(b)). We report that this individual parameterization is more robust to outliers and capable of capturing features that cannot be detected by standard methods (e.g. 'scissors' example in Figure \ref{fig:standard})
    

    

    \item[\textbf{3.}] We demonstrate that the computation of the GPS is scalable. This is  because the parameter generator for the GPS can be computed by standard stochastic optimization methods. 
    

\end{enumerate}

\section{Predictive Matching Generative Parameter Sampler}
\label{sec:gps}
The main idea of GPS abstracts the notion of an \emph{Integral Probability Metric} (IPM) \citep{muller1997integral} to measure a distance between the predictive distribution and its empirical counterpart. Some examples of IPM include Wasssestein distance \citep{villani2008optimal}, energy distance \citep{szekely2013energy}, and \emph{Maximum Mean Discrepancy} (MMD) \citep{gretton2012kernel}. For computational convenience, we use MMD as a default choice of the distance in this article. 

 For $i=1,\dots,n$, we posit that our observations $y_i\mid\theta_i$ are generated from a density function $f_{\theta_i}$ with parameters $\theta_i\in\mathbb{R}^p$, independently for $i=1,\dots,n$. We define a random variable $Z_i$ that can be easily generated (such as $Z_i\sim N(0,I_q)$), and pass it through a function $G:\mathbb{R}^{q}\rightarrow\mathbb{R}^{p}$ to generate a parameter $\theta_i:=G(Z_i)$. We call $G$ the \emph{generator}, and we denote the  density function of $\theta_i$ induced from the generator $G$ by $\pi_G$. The predictive distribution for $\pi_G$ can thus be written succinctly as
\begin{align}\label{eq:pred}
\widetilde f_{G}(y_i):=\int f_{\theta_i}(y_i)\pi_G (d\theta_i).
\end{align}
 We model the generator $G$ by a neural network and we denote the  parameter of the generator by $\phi$. The generator function evaluated by parameter $\phi$ is denoted by $G_\phi$. We proceed to give a definition of the optimal $\phi$ as below:
\begin{definition}
The optimal generator of GPS is defined as the minimizer of the MMD between the predictive distribution $\widetilde f_{G_\phi}$ and the empirical distribution $f_n$ :
\begin{eqnarray}\label{eq:GPS}
\widehat \phi = \argmin_\phi \emph{\textrm{MMD}}(f_n, \widetilde f_{G_\phi}).
\end{eqnarray}
Following that, the UQ distribution is defined by the distribution of $G_{\widehat\phi}(Z)$, where $Z\sim N(0,I_q)$.
\end{definition}

\begin{figure}[t!]
\centering
\begin{subfigure}{.5\textwidth}
  \centering
  \includegraphics[width=.45\linewidth]{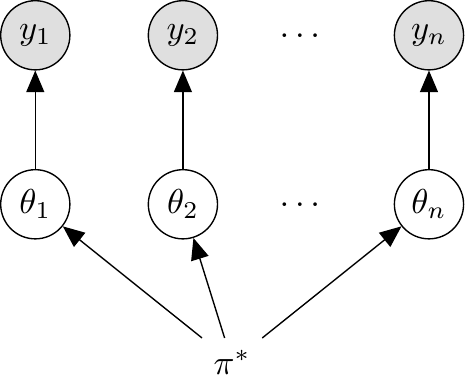}
  \caption{\footnotesize  Individual parameterization}
  \label{fig:sub1}
\end{subfigure}%
\begin{subfigure}{.5\textwidth}
  \centering
  \includegraphics[width=.45 \linewidth]{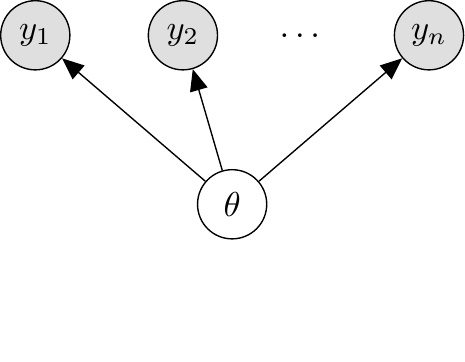}
  \caption{\footnotesize  Global parameterization (Bayesian and frequentist)}
  \label{fig:sub2}
\end{subfigure}
\caption{\footnotesize  Parameterization of GPS. }
\label{fig:gps-vs-standard}
\end{figure}
 
For convenience, we consider the negation of MMD here. Once the generator $G_{\widehat{\phi}}$  is trained, the parameters can be generated from the UQ distribution by the following procedure: i) sample $Z^{(\ell)}\sim N(0,I_q)$; ii) evaluate $\theta^{(\ell)}=G_{\widehat{\phi}}(Z^{(\ell)})$ for $\ell=1,\dots,N$, where $N$ is some large-sized integer. We then use the sampled parameters $\theta^{(1)},\dots,\theta^{(N)}$ to derive the characteristics of the UQ distribution such as the mean or  marginal 95\% uncertainty interval, say. The UQ for prediction can be also easily implemented by sampling $\widetilde y_{pred}^{(\ell)}\sim f_{\theta^{(\ell)}}$ for $\ell=1,\dots,N$, and a prediction interval can be evaluated from the empirical distribution of the sampled predictions. 

\noindent{\bf Remark.} The core of Bayesian procedures emphasizes the \emph{posterior} that is defined as a conditional distribution of the parameter given the observations. On the other hand, while our parameter is modeled by a distribution (as in the Bayesian setting), the UQ distribution of the GPS does not follow a conditional law, which distinguishes the GPS from the standard Bayesian framework.



\noindent{\bf Individual parameterization for GPS.} \emph{Individual parametrization} indicates that each $y_i$ is related to a different $\theta_i\in\mathbb{R}^p$ (confer Figure  \ref{fig:sub1}). The individual parameterization in the GPS assumes that there exists a true parameter-generating law $\pi^*$ for each $\theta_i$ for $i=1,\dots,n$. This setup can be considered as a slight generalization of the global parameterization 
(Figure \ref{fig:sub2}). This is because  placing a point mass $\pi^*$ on a fixed $\theta_0$ results in a standard frequentist model.

The main reason of constructing the individual parameterization is to exploit a scalable computation under the \emph {fully} independent structure between the observations. This means that any set of generated predictive samples can be used to approximate the MMD in optimization. In contrast, under the classical setup as in Figure \ref{fig:sub2}, the computation of a distance in GPS requires a generation of  a full set ($n$-sized) of predictive samples to completely quantify the conditional reasoning $\theta\mid y_1,\dots, y_n$. This standard Bayesian models are  computationally demanding for large-sized data sets.     


The individual parameterization has not been commonly used in practice. In Bayesian statistics, Dirichlet Process Mixture (DPM) models follow the individual parameterization by imposing a DPM prior on the parameters \citep{teh2005sharing,teh2010dirichlet,maceachern1998estimating}, but the MCMC computation of the posterior distribution is extremely demanding especially for large-sized data sets. This is because  every iteration the MCMC algorithm samples $n$ number of $\theta$s. Moreover, the DPM models have a nature of discreteness in the posterior distribution. While its discreteness property is helpful in  clustering problems without specifying the number of clusters \emph{a priori}, it hurdles the practical use of the DPM to more general settings where the posterior space is continuous. For these reasons, we do not examine the DPM models further in sequel.

In a frequentist paradigm, the individual parameterization is not considered because it is assumed that the true parameter is fixed without any randomness on the parameter.
While some hierarchical models such as random-effects models are considered in practice, its main focus is not on estimating the distribution of $\theta$, but in reducing some random-effects that may affect the estimation of parameters of interest.

The well-known Gaussian sequence model, $y_i=\theta_i+\epsilon_i$, also inherits the individual parmeterization setup. However, this model is not  designed to analyze real data sets, but its usage is mainly focused on investigating theoretical properties of various statistical procedures. These include risk minimization \citep{stein1981estimation, johnstone2004needles, castillo2012needles}, high-dimensional variable selection \citep{castillo2015bayesian, rovckova2018bayesian} and nonparametric function estimation \citep{johnstone1999wavelets}. For this reason, we also do not discuss the Gaussian sequence model further.      

In Section \ref{sec:application}, we present a thorough investigation on the advantages of using the  individual parameterization setup.  

\noindent{\bf Objective function and computation.} Using the definition of MMD \citep{gretton2012kernel}, our objective function for GPS takes the form 
\begin{eqnarray}
\label{eq:MMD}
\mbox{MMD}(f_n, \widetilde f_G) = -\frac{2}{n}\sum_{i=1}^n\mathbb{E}_{\widetilde Y_i\sim \widetilde f_{G}}[k(y_i,  \widetilde Y_{i} )  ] + \frac{1}{n}\sum_{i=1}^n\mathbb{E}_{\widetilde{Y}_i,\widetilde{Y}_i'\overset{i.i.d.}{\sim}\widetilde f_{G}}[k (\widetilde Y_{i}  , \widetilde Y^{\prime}_{i} )  ]
\end{eqnarray}
where $k(\cdot,\cdot)$ is a positive-definite kernel. In this article, we use a Gaussian kernel $k(x_1,x_2) = \exp\{-\Vert x_1 - x_2\Vert^2_2\}$ as a canonical choice. In practice, any valid kernels may be used as well. Algorithm \ref{alg:alg} demonstrates a \emph{Stochastic Gradient Descent} (SGD) algorithm to minimize the objective function \eqref{eq:MMD}. The main idea of this algorithm is to approximate the expectation in \eqref{eq:MMD} by a Monte Carlo (MC) procedure.  At every iteration, its MC approximation is conducted by sampling different parameters accompanied by their  predictive samples. The number of samples for the MC approximation can be practically small, such as $M=10$ and $J=5$, say. We show that this setting achieves superior empirical performance in Section \ref{sec:exp}. While Algorithm \ref{alg:alg} adopts an SGD procedure, other stochastic optimization procedures, such as \texttt{Adagrad} \citep{duchi2011adaptive} and \texttt{Adam} \citep{kingma2014adam}, can be deployed to accelerate the convergence. 


\begin{algorithm}
\footnotesize
\caption{\footnotesize  An SGD algorithm for the GPS.}\label{alg:alg}
\begin{algorithmic}
\State Set the learning rate $\gamma_0$, $n_0$, $M$, $J$, and set $t=0$
\While{the stop condition is not satisfied} 
\State{Set $J_{\phi^{(t)}}=0$.}
\State{Sub-sample $n_0$ samples from the original data set.}
\For{$m = 1,\dots,M$}
\For{$i$ in the sub-sampled indexes}
\State{Independently generate $z_i^{(m)}$ from $N(0,I_q)$.}
\State{Set $\theta_i^{(m,t)}=G_{\phi^{(t)}}(z_i^{(m)})$.}
\For{$j=1,\dots,J$}
\State{Independently generate $\widetilde y_{ G_{\phi^{(t)}} ,i(j) }^{(m)}$ from $f_{ \theta_i^{(m,t)} }$.}
\EndFor
\EndFor
\State{Set $J_{G_{\phi^{(t)}}} = J_{G_{\phi^{(t)}}} + \nabla_\phi\widehat{d}(f_{n_0},\widetilde f_{G_{\phi}})\big\vert_{\phi = \phi^{(t)}}$, where $f_{n_0}$ is  the empirical density of the sub-sampled data points and $\widehat d$ is a Monte Carlo approximation of MMD in \eqref{eq:MMD} from sampled  $\widetilde y_{G_{\phi^{(t)}},i(j)}^{(m)}$ for $j=1,\dots,J$.  }
\EndFor
\State{$\phi^{(t+1)} = \phi^{(t)} - \gamma_t J_{\phi^{(t)}}$.} 
\State{Set $t = t+1$.}
\EndWhile
\end{algorithmic}
\end{algorithm}

\section{Applications}\label{sec:application}
       
\subsection{ Linear regression.}\label{sec:linear}
Consider a linear regression model that contains a univariate response variable $y_i$ with a $p$-dimensional predictor $X_i$ for $i=1\dots,n$. Its regression coefficient is denoted by $\theta\in\mathbb{R}^p$; i.e., $ y_i = X_i^T\theta + \epsilon_i$,
where $\epsilon_i\sim N(0,\sigma^2)$. In the framework of  GPS, we consider an individual parameterization such that the uncertainty on  $\{\theta_i,\sigma^2_i\}_{i=1,\dots,n}$ is quantified by the UQ distribution. We let $G(Z_i)=\{\theta_i,\sigma^2_i\}$.

\begin{figure}
    \centering
        \includegraphics[width=0.7\textwidth]{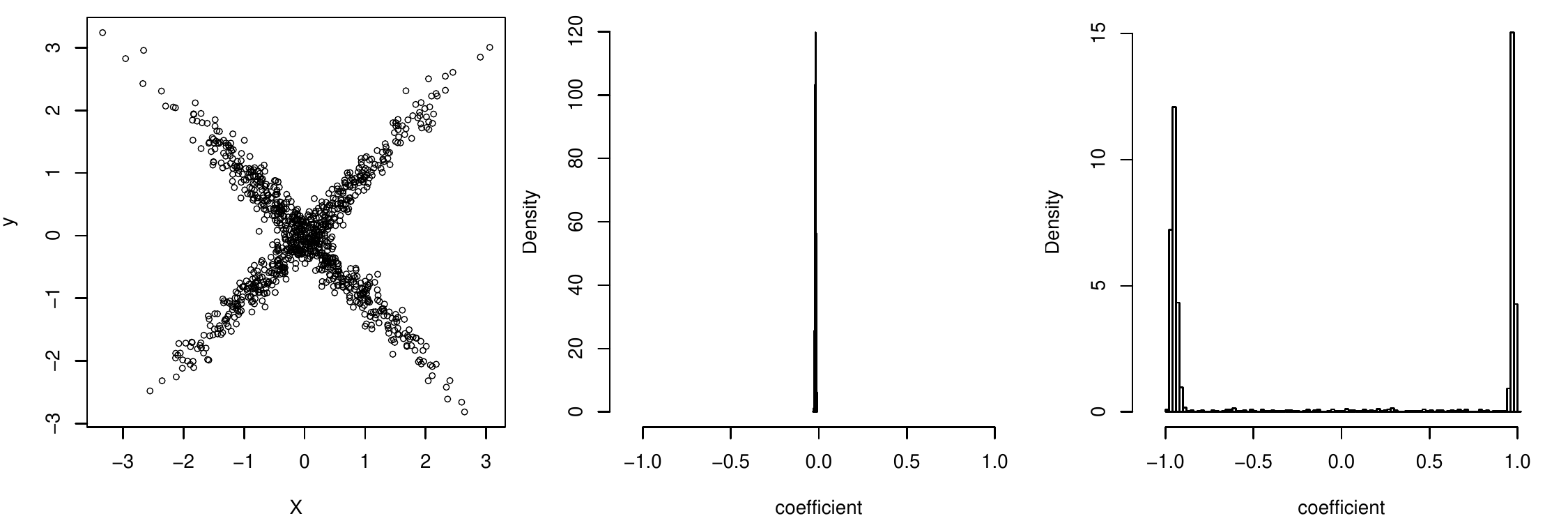}
       \caption{\footnotesize  The scatter plot of a scissors example (left); Bayesian posterior distribution of $\theta$ (middle); the UQ distribution from the GPS (right).} \label{fig:standard}
\end{figure}

We first consider the case of $p=1$ and $n=2000$, where $X_i$ is generated from i.i.d. standard Gaussian. The true data-generating process follows $y_i=\gamma_iX_i - (1-\gamma_i)X_i  +\epsilon_i$, where $\gamma_i\sim \texttt{Bernoulli}(1/2)$ and $\epsilon_i\sim N(0,0.2^2)$ for $i=1,\dots,n$. 

The resulting scatter plot of a synthetic data set is illustrated in Figure \ref{fig:standard} (left). In  this example, the classical global parameterization is not capable of capturing the `scissors' like shape. As a consequence, we see from Figure \ref{fig:standard} (middle) that the standard posterior distribution of $\theta$  is  concentrated on zero, when a uniform prior is used, which is completely different from the true nature.  In contrast, Figure \ref{fig:standard} (right) shows that the UQ distribution derived from the GPS captures exactly  the bimodal shape of the true distribution on the slope parameter ($\theta_i=1$ with $50\%$ and $-1$ with $50\%$). This example shows that compared to standard procedures, the GPS covers a wider range of true data-generating processes.

To examine the robustness of the GPS to outliers, we consider an example as follows: we generate $X_i\in\mathbb{R}^{20}$ i.i.d. from  $N(0,\Sigma)$, where $\Sigma_{jk} = 0.5$, if $j\neq k$ and $\Sigma_{jj} =1$ for $j,k=1,\dots,20$, $\epsilon_i\sim N(0,1)$, and $n=200$. The 5\% of response samples are randomly set to be outliers and they are contaminated with extra noises generated from a Cauchy distribution with a scale value of $3$. 

\begin{figure}
    \centering
        \includegraphics[width=3.5in, height = 1.2in]{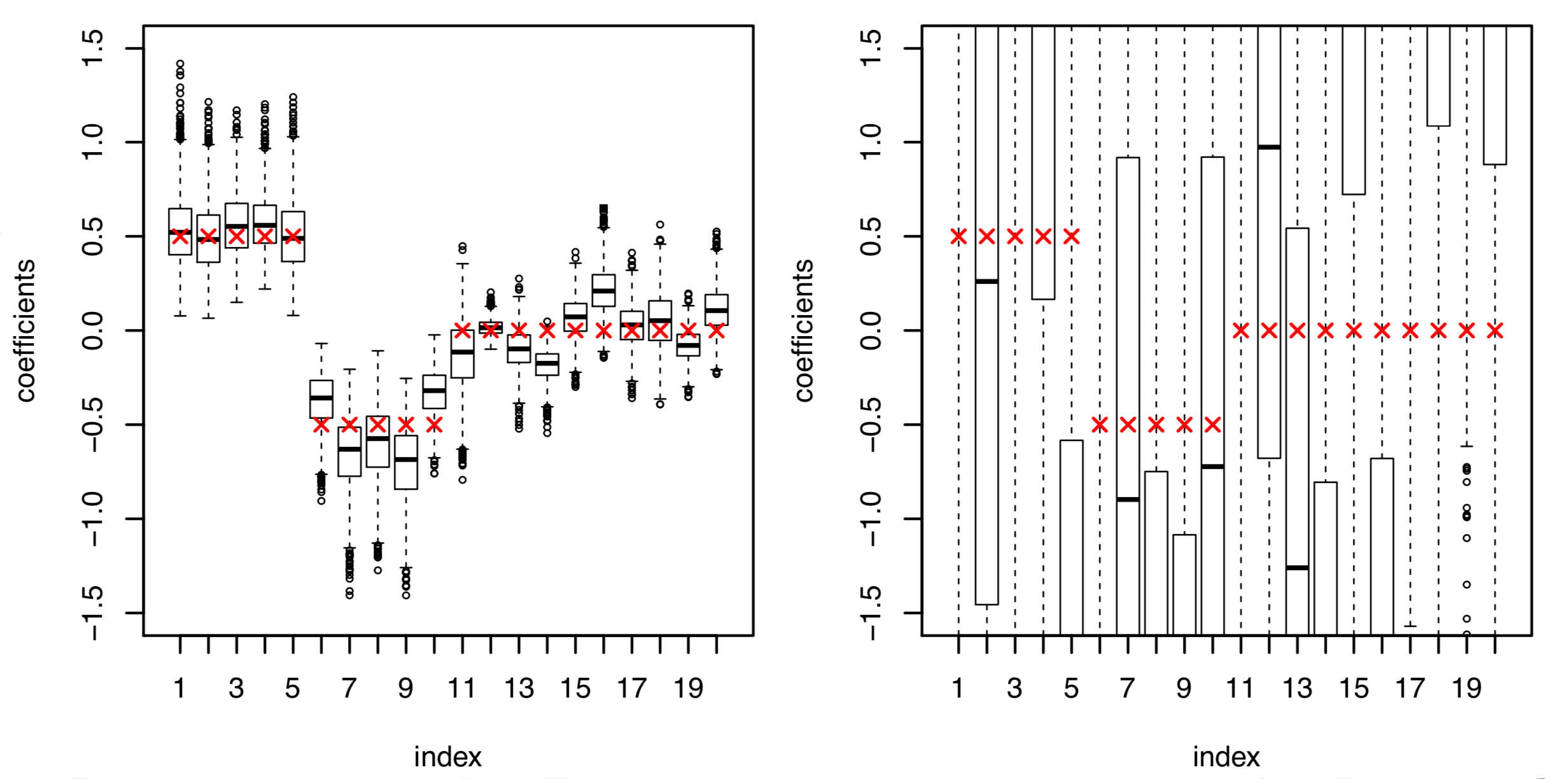}
        \includegraphics[width=3.5in, height = 1.2in]{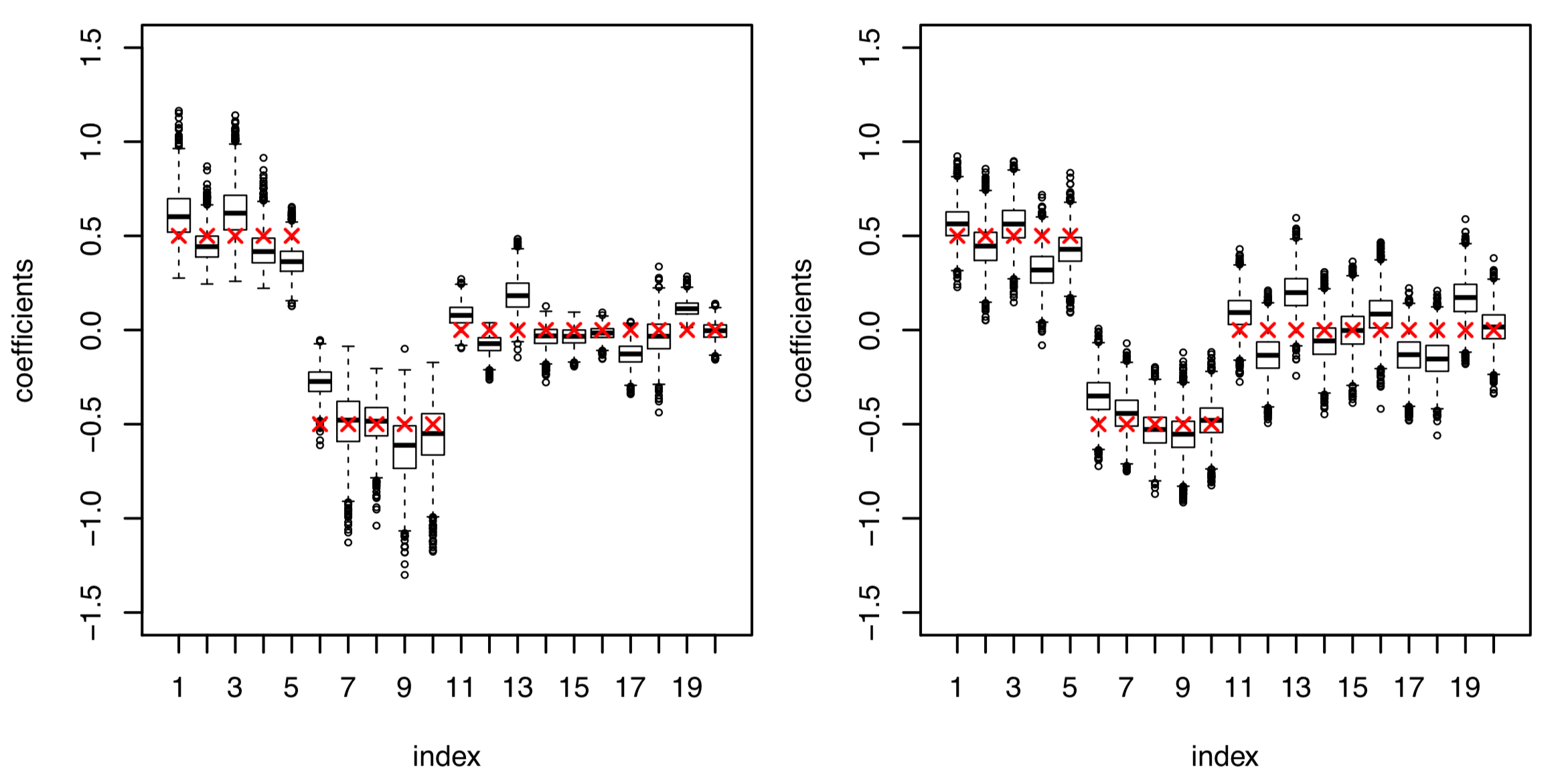}
       \caption{\footnotesize An example under the existence of outliers (top) and  under the absence of  outliers (bottom) for the UQ distribution of the GPS (left), the Bayesian posterior distribution (right). The true coefficient values are marked by red ``\texttt{x}''. } \label{fig:outliers}
\end{figure}

 We compare the UQ distribution to the Bayesian posterior distribution (with a uniform prior) as illustrated in the top row of Figure \ref{fig:outliers}. While the existence of outliers distorts the posterior behavior, the resulting UQ distribution of GPS is reasonably concentrated around the true parameters. This show that GPS is more robust to outliers than the standard Bayesian procedure in this simple example. 

These results present that the GPS is robust to outliers as well as being able to recover non-standard features (`scissors' shape). These desirable properties stem from a characteristic of the individual parameterization such that  each observation is only affected from an individual parameter without interfering the other parameters. This explains why the GPS is robust to outliers. Even if there exist a small number of outliers, under the individual parameterization the effect of the outliers is minimal to the individual parameters corresponding to non-outliers.    

We also consider a classical scenario where the  assumptions for the linear model are fully satisfied. We examine the same setting used in the previous `outlier' example, but this classical example contains no outliers. The results are illustrated in the bottom row of Figure \ref{fig:outliers}. The UQ distribution behaves similarly with the Bayesian posterior distribution. 


\subsection{\bf Poisson processes with random intensity.} We consider an inhomogeneous Poisson process on a spatial domain $\mathcal{S}\in\mathbb{R}^2$ which is parameterized by an intensity function $\lambda(s):\mathcal{S}\rightarrow R^{+}$ \citep{kingman1993poisson}. The random number of events in region $\mathcal{T}\subset\mathcal{S}$ is Poisson random variable with parameter $\lambda_{\mathcal{T}}=\int_{\mathcal{T}}\lambda(s)ds$. 





\begin{figure}
    \centering
        \includegraphics[width= 0.49\textwidth]{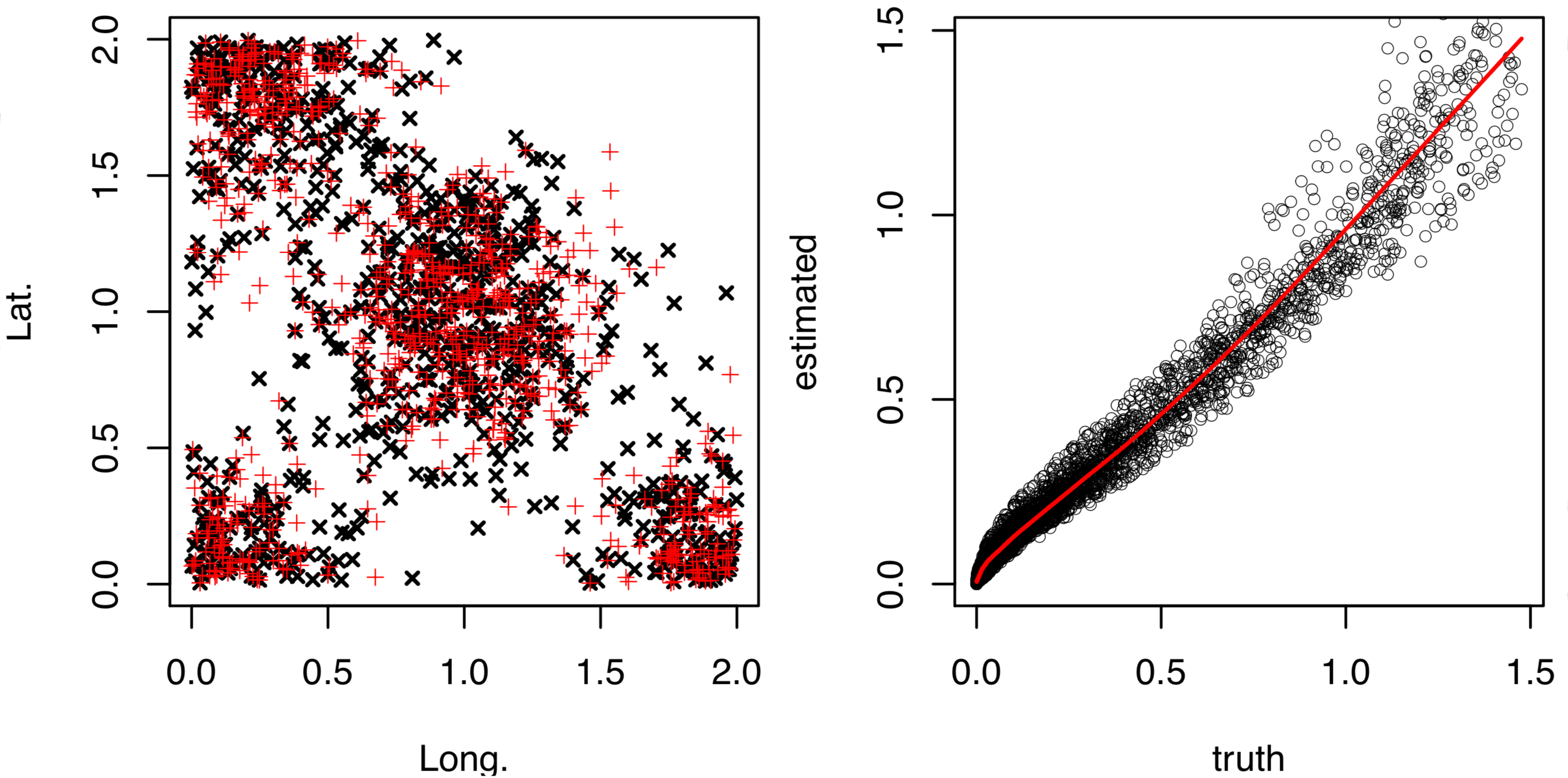}
       \includegraphics[width= 0.49\textwidth]{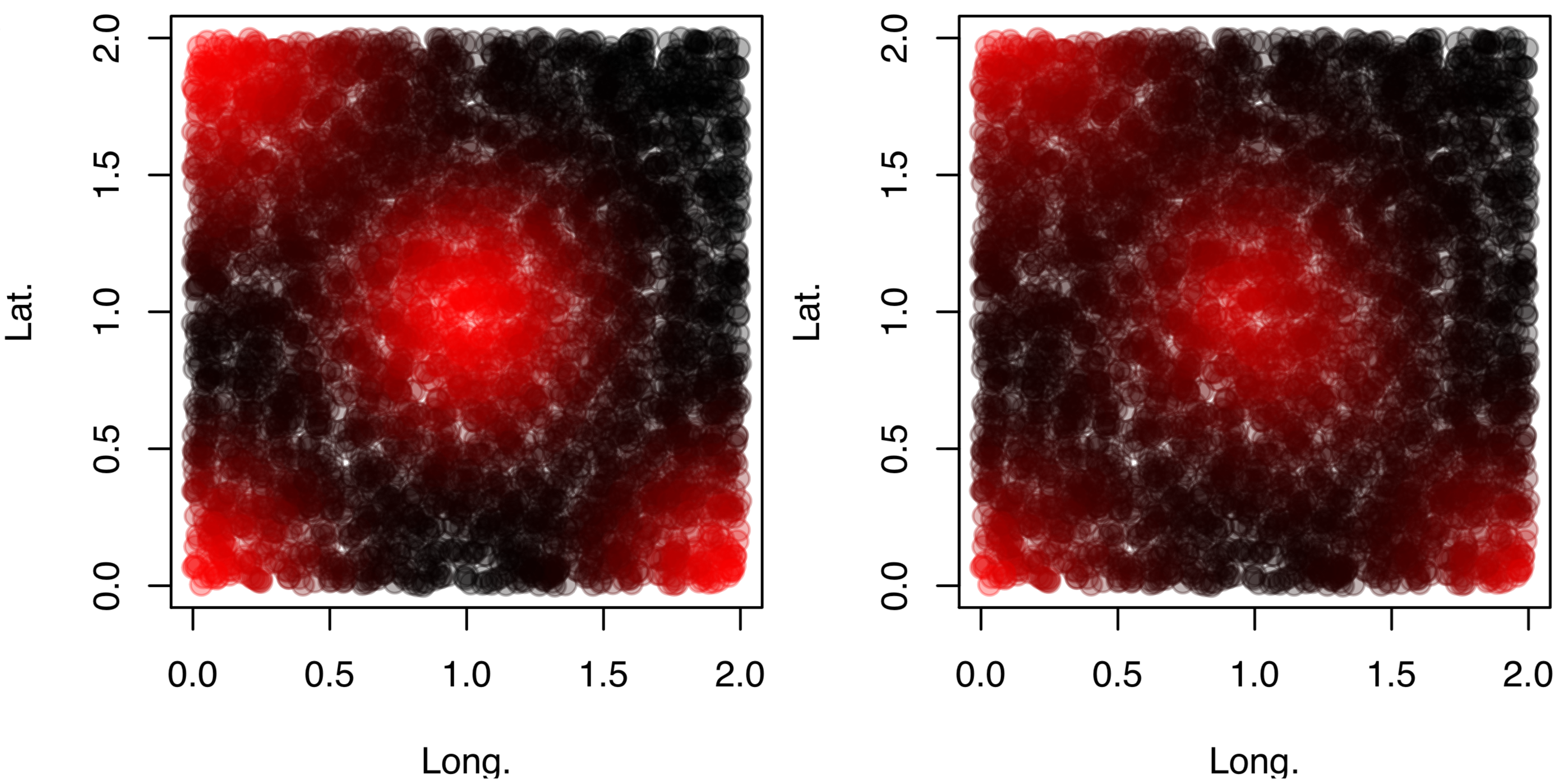}
       \caption{\footnotesize   Generated samples by the GPS (black ``\texttt{x}'') and the observations (red ``\texttt{+}'') (first); the mean of the intensity evaluated by the GPS versus the truth (second); the true intensity on the spatial domain (third); the mean of intensity by the GPS on the spatial domain (fourth). } \label{fig:poisson}
\end{figure}

The GPS  is tested on a  synthetic data set generated with the true intensity function. This intensity function is
illustrated as a heatmap in the third plot of Figure \ref{fig:poisson}.  To embed this in our GPS framework, we let $\theta_i:=\lambda(s_i)$ and $\log \theta_i=G_\phi(\{s_i,Z_i\})$, where $s_i\in\mathbb{R}^2$. We then train the generator which is modulating the intensity function $\lambda(\cdot)$. From the learned intensity through the GPS, we generate the points via thinning \citep{Lewis:1978}. The first plot  of Figure \ref{fig:poisson} illustrates one path of the simulated counts compared with the realized observations. The fourth figure depicts the mean of intensity over $1,000$ samples drawn from the GPS. Overall, this confirms that GPS operates sensibly in that it is able to recover the ground truth.

\subsection{Deep neural network for classification}\label{sec:DNN}
In image classification, Deep Neural Network (DNN) models have been widely used by deploying convolutional layers \citep{krizhevsky2012imagenet}, while the uncertainty quantification on the prediction of the classification still remains a challenging task. To solve this uncertainty quantification problem, we apply the GPS to the classification model with $H$ number of classes that follows: for $h=1,\dots,H$, $\mathbb{P}[y_i=h] = \exp\{\theta_{hi}\}/(\sum_{u=1}^H\exp\{\theta_{ui}\})$, where $X_i\in\mathbb{R}^p$ is the the $i$-th predictor (or image) corresponding to $y_i$, and $\{\theta_{1i},\dots,\theta_{Hi}\}$ denotes the classifier of $X_i$ for each class. The classifier function  is commonly modeled by a DNN (e.g., Convolutional Neural Network (CNN)) by imposing $X_i$ as the input and $\{\theta_{1i},\dots,\theta_{Hi}\}$ as the output, and the cross entropy loss can be used to train the DNN in standard procedures. 

\begin{figure}
\centering
  \includegraphics[width=4.6in]{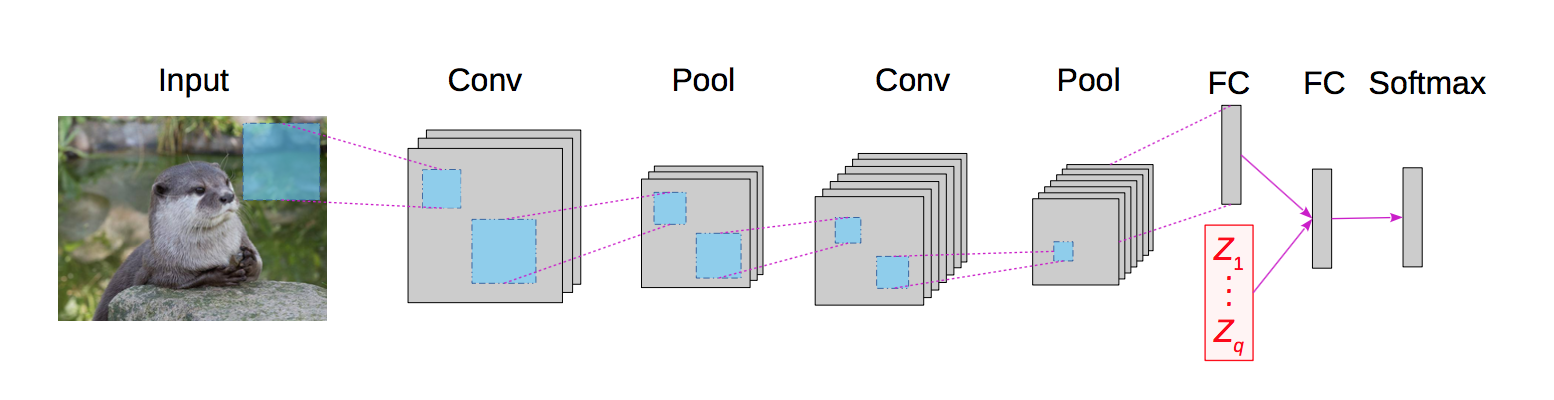}
  \caption{\footnotesize An example of the GPS for classification tasks using a CNN. }
  \label{fig:cnn}
\end{figure}

 For this application, we set the GPS as $\{\theta_{1i},\dots,\theta_{Hi}\} = G_\phi(\{X_i, Z_i\})$ such that $G_\phi: \mathbb{R}^{p+q}\to\mathbb{R}^H$, where  $\phi$ denotes the parameters of  the generator and $q$,  the dimension of $Z_i\sim N(0, I_q)$. The only difference from this GPS setting and the standard DNN structures is that this generator embraces the randomness to the neural network by augmenting the random noise variable $Z_i$ into the input, so that the resulting neural network is naturally random (see Figure \ref{fig:cnn}). After learning the parameters $\phi$ for the generator by optimizing the objective function in \eqref{eq:MMD}, the trained generator of the classifiers then generates the classification probabilities that captures the uncertainty in matching the predictions and the observations. The CNN based on GPS can be trained by implementing  Algorithm \ref{alg:alg}.

This GPS structure in Figure \ref{fig:cnn} also enjoys a computational advantage over the other uncertainty quantification procedures based on variational inference. These include inverse autoregressive flow \citep{kingma2016improved}, non-linear independent components estimation \citep{dinh2014nice} and normalizing flow \citep{Rezende:2015}. Like the GPS, these procedures also consider an idea of generator using a noise random variable. However, unlike the GPS, these procedures approximate the Bayesian posterior distribution of the DNN (or CNN) parameter that is usually extremely high-dimensional. Constructing a neural network that transforms a random noise $Z\sim N(0,I_q)$ to such a high dimension (the number of parameters in the target DNN) is a computationally challenging task. For example, it may require more than tens of billions of  generator parameters to be trained, when the target dimension is a few millions. In contrast, the GPS directly models the classifier function by using a DNN augmenting the random noise $Z$ into the input. The only difference from the deterministic DNN is that the input dimension is increased just by the dimension of $Z$. This dimension can be controlled by the user. In Section \ref{sec:exp}, we show that the performance of the generator is not sensitive to the choice of the dimension of $Z$ in some examples.

In addition to the problem of high-dimensionality, the variational inference procedures also require a strict restriction on the network structure of  the generator. This is to ease the computation of the determinant of the resulting Jacobian term. This restriction sometimes may slow down the convergence of training, because it limits the flexibility of the neural network. In contrast, the GPS bypasses the calculation of the Jacobian term by adopting the MMD as the distance measure. 




\section{Experiments on MNIST and CiFAR10 Data Sets}\label{sec:exp}
In this section, we evaluate the performance of the GPS  on classification tasks and uncertainty quantification using MNIST \citep{lecun1998mnist} and CiFAR10 \citep{krizhevsky2009learning} data sets. 




\noindent{\bf Uncertain images in classification.}
We first describe the meaning of ``Uncertain'' images in the classification model based on the GPS. Intuitively, uncertain images are images that the classification probabilities induced by the UQ distribution have large probabilistic fluctuations so that none of the classification probabilities dominate the other classes.  

\begin{figure}
    \centering
        \begin{subfigure}{.5\textwidth}
  \centering
          \includegraphics[width=2.3in, height = 1.1in]{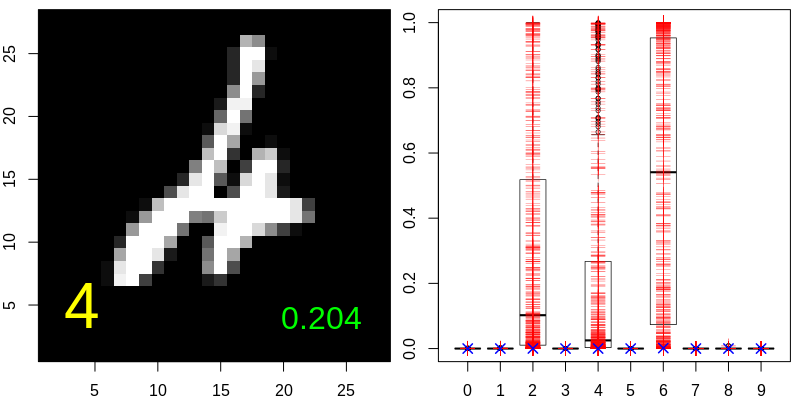}
  \caption{An example of ``Uncertain'' images}
  \label{fig:uncertain0}
\end{subfigure}%
\begin{subfigure}{.5\textwidth}
  \centering
          \includegraphics[width=2.3in, height = 1.1in]{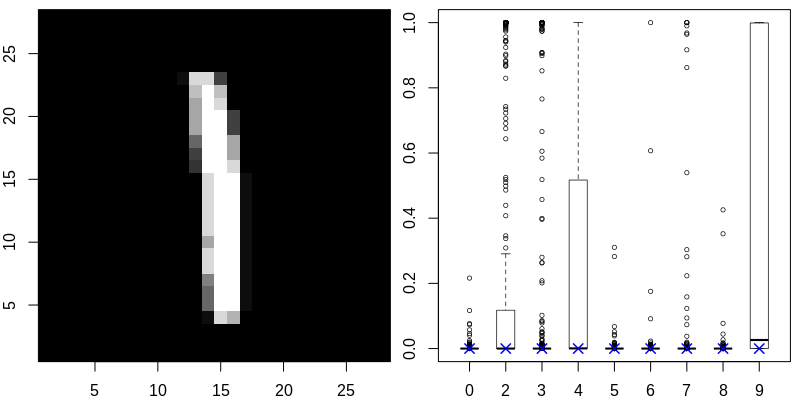}
  \caption{An example where the digit ``1'' is ignored in the training}
  \label{fig:without0}
\end{subfigure}

       \caption{\footnotesize The yellow  number: the the true label; the green value: the mean of classification probability  of the true class. The boxplots illustrate the UQ distributions of the classification probabilities and the red crosses in the boxplot show the sampled classification probabilities from the UQ distribution. The UQC in \eqref{eq:t_h} for each class is marked by blue ``\texttt{x}''. } 
\end{figure}

In order to carry out our experiment, we formally  define those images that are uncertain in the following manner. For each class $h$ and an image $X$, we propose an \emph{Uncertainty Quantification Criterion} (UQC), denoted by $t_h$ such that
\begin{eqnarray}\label{eq:t_h}
    \mathbb{P}_Z\left[ C_h(\{X,Z\})< t_h \right] = 5\%,
\end{eqnarray}
where $C_h(\{X,Z\})$ is the classification probability of the class $h$, i.e., $C_h(\{X,Z\})= \exp\{G_h(\{X,Z\})\}/\sum_{l=1}^H \exp\{G_l(\{X,Z\})\}$ for $h=1,\dots,H$. The value of $t_h$ can easily be  approximated by generating classification probabilities via the trained UQ distribution. We call an image ``Uncertain'' if  every $t_h$ corresponding to the image does not exceed 50\%. This means that the classification probabilities of an ``Uncertain'' image are not dominated by a single class.  If an image is not ``Uncertain'', we call it ``Certain''. An example of ``Uncertain'' images in the MNIST data set is illustrated in Figure \ref{fig:uncertain0}, and the distribution of marginal classification probabilities for digits ``2'', ``4'', and ``6'' are too disperse so that none of classification probabilities satisfies the condition of certainty.  

The categorization of ``Uncertain'' images is practically useful in a sense that the model uncertainty examines whether the model is certain about  the prediction on a given data set. This uncertainty quantification procedure enables us to avoid an overconfident decision under an existence of high uncertainty for prediction. In the following, we show that the error rate of classification can be dramatically reduced by ignoring ``Uncertain'' images.


\noindent{\bf Classification under uncertainty quantification.} Here we use a CNN \citep{krizhevsky2012imagenet} for the MNIST and a ResNet \citep{he2016deep} for the CiFAR10 data set (the detailed settings are listed in the Appendix).  For all procedures, we consider $100$ epochs and $1,000$ epochs in optimization for the MNIST and the CiFAR10 in training, respectively. The mini-batch size  is set to be $100$ for both data sets.  We set the dimension of $Z$ to be $q=r\times p$, where $p$ is the original input dimension to the feed-forwarding network for $r=0.5, 0.25$. The classification tasks for the GPS are implemented by using the mean of the classification probabilities based on the UQ distribution, i.e., $\mathbb{E}_Z\left [ C_h(\{X,Z\}) \right]$ for $h=1,\dots,H$. To train neural networks, we use Algorithm \ref{alg:alg} with $\gamma_t=0.01\cdot t^{-1/2}$, $M=10$ and $J=5$, and its \texttt{pytorch} code is available in \url{https://github.com/minsuk000/GPS}.

To examine the quality of the uncertainty quantification by the GPS, we compare our GPS to the Monte Carlo dropout \citep{gal2016dropout} (MCDrop in short). The MCDrop procedure is implemented by using a dropout \citep{srivastava2014dropout} during training of the neural network. Then, the output of the trained network is evaluated by randomly dropping the nodes in the network, so the resulting output is random. \cite{gal2016dropout} showed that the MCDrop is an approximation of  the variational predictive  posterior distribution that minimizes the KL-divergence towards a deep Gaussian process \citep{damianou2013deep}. Due to the stochastic nature in the MCDrop procedure, we can also  evaluate the UQC as in the GPS, and ``Uncertain'' images can be determined by  the UQC derived from the MCDrop.       

\begin{table}[htbp]
\caption{\footnotesize Test error rates under uncertainty quantification. \label{tab:results}
}
\centering
\resizebox{0.8\columnwidth}{!}{
\begin{tabular}{l|ccc} 
\hline
\multicolumn{4}{c}{{\bf MNIST} } \\
\hline
Method & Error Rate (Full)& Error Rate (w\slash o ``Uncertain'' )& \% of ``Uncertain''\\
\hline
Standard CNN & 0.79\% &  &      \\
CNN-MCDrop & 0.80\% & 0.30\% & 2.05\% \\
CNN-GPS ($r=0.5$) & 0.72\% & {\bf 0.12}\% & 2.99\%     \\
CNN-GPS ($r=0.25$) & {\bf 0.69}\% & 0.15\% & 2.27\%   \\
\hline
\hline
\multicolumn{4}{c}{{\bf CiFAR10} } \\
\hline 
Standard ResNet & 16.50\% & &  \\
ResNet-MCDrop & 16.65\% & 10.37\% & 13.83\% \\
ResNet-GPS ($r=0.5$) & {\bf 15.95}\% & {\bf 6.811}\% & 22.48\% \\
ResNet-GPS ($r=0.25$) & 17.36\% & 7.892\% & 24.54\% \\

\hline 
\end{tabular}
\footnotesize }
\end{table}

\begin{table}[htbp]
\caption{\footnotesize  Trained without one class. 
}\label{tab:uncertain}
\centering
\resizebox{0.8\columnwidth}{!}{
\begin{tabular}{l|cc} 
\hline
\multicolumn{3}{c}{{\bf MNIST} } \\
\hline
Method & \% of ``Uncertain'' from unseen class & \% of ``Uncertain'' from random noise \\
\hline
CNN-MCDrop & 68.90\% & 69.76\% \\
CNN-GPS  & 97.27\%  & 97.28\% \\
\hline
\hline
\multicolumn{3}{c}{{\bf CiFAR10} } \\
\hline
ResNet-MCDrop & 53.50\% & 47.53\%  \\
ResNet-GPS  & 72.30\% & 77.24\% \\
\hline

\hline
\end{tabular}
}
\end{table}

In Table \ref{tab:results}, we report the classification performance of the standard procedure and our GPS procedure. More precisely, we provide the error rate of classification from the full test data set, the proportion of ``Uncertain'' images, and the error rate from the test data set excluding the ``Uncertain'' images. The GPS procedure achieves competitive classification performance for both MNIST and CiFAR10 data sets. After discarding ``Uncertain'' images, the classification accuracy is dramatically improved for the GPS procedure. Especially for the MNIST data set, only less than 3\% of images are discarded, but the error rate is decreased by a factor of six; from $0.72\%$ to $0.12\%$ ($r=0.5$) and from $0.69\%$ to $0.15\%$ ($r=0.25$). These results outperform the state-of-art error rate for the MNIST data set ($0.21\%$; \cite{wan2013regularization}). This improvement means that the UQ derived by the GPS is truly capable of capturing uncertain situations and helping  practitioners to make a better decision. In contrast, even though the proportion of ``Uncertain'' images determined by the MCDrop is slightly less than that by the GPS, the error rate on its ``Certain'' images is more than twice of these from the GPS.


Overconfident predictions on unseen classes in classifications are problematic and calls for caution. In an ideal situation, we would like the GPS to detect higher uncertainty when the test data significantly differs from the training data. To check if the GPS inherits this desirable property, we consider an extra experiment where we remove one class of images (digits ``1'' for MNIST and class ``truck'' for CiFAR10) in training steps. Then, we compute the  classification probabilities evaluated from the unused images and random noise generated from i.i.d uniform distribution from $0$ and $1$. We use $r=0.5$ for the GPS.

Figure \ref{fig:without0} presents an example of  images with digit ``1'' and their classification probabilities under a setting described in the previous paragraph. Because the digit ``1'' has not been used in the training steps, it is reasonable that the classification probability of each class is not close to unity and the resulting distribution should be disperse. The value of $t_h$ for each class $h$ is close to zero and classification probabilities of some classes are widely distributed  in the range of $0$ and $1$.

Table \ref{tab:uncertain} contains the proportion of ``Uncertain'' images from an unseen class during training steps for the MNIST  and the CiFAR10 data set. In the MNIST example, the GPS procedure detects images that are significantly different from the training data set  with high chance ($97.27\%$ for digit ``1'' and $97.28\%$ for random noise), while the MCDrop determines only less than 70\% images to ``Uncertain''. In the CiFAR10, the detection rate of the GPS is $72.30\%$ for class ``truck'' and $77.24\%$ for random noise. In contrast, the  rate of the MCDrop is around 50\%.  



\section{Conclusion }
We proposed a general framework, called GPS, that is computationally scalable in quantifying uncertainty in estimating parameters. We showed that the GPS can be applied to a wide range of models, e.g., linear models, Poisson processes, and deep learning. With experiments carried out, we conclude that the GPS is successful in providing a model-based  uncertainty  quantification.



\bibliographystyle{named}
\bibliography{ref}
\newpage
\appendix
\section*{Appendix}
\noindent{\bf Detailed Neural Net Settings.} In this section, we provide the detailed settings of the  deep convolutional neural networks for the MNIST data set and the CiFAR10 data set used in Section \ref{sec:exp}. 

We adopt the notation that \texttt{conv}$[N,w,s,p]$ denotes a convolutional layer with $N$ filters of size $w\times w$, with stride $s$ and $p$ pixel padding, and \texttt{max-pool}$[s]$ is a $s\times s$ max-pooling layer with stride $s$, and \texttt{BN} denotes a batch normalization \citep{ioffe2015batch} step. We denote an average-pooling layer with stride $s$ by \texttt{avg-pool}. Also the ReLU function, $\sigma(t) = \max\{0,t\}$, is denoted by \texttt{ReLU}, and \texttt{convres}$[N,w,s,p]$ denotes a convolutional layer \texttt{conv}$[N,w,s,p]$ with an addition to an identity function, used in the ResNet \citep{he2016deep}.  

The CNN used for the MNIST data set follows as
\texttt{conv}$[16,5,1,2]$-\texttt{BN}-\texttt{ReLU}-\texttt{max-pool}$[2]$-\texttt{conv}$[32,5,1,2]$-\texttt{BN}-\texttt{ReLU}-\texttt{max-pool}$[2]$, and the flattened output of this convolutional layer (size $7\times7\times32$) is connected to a 6-layered feed-forwarding neural network. Each layer in the feed-forwarding network contains the same number of nodes of $800$, and each node in the feed-forwarding network is batch-normalized. 

The ResNet for the CiFAR data set follows 
\texttt{conv}$[16,3,1,1]$-\texttt{BN}-\texttt{ReLU}-\texttt{convres}$[16,3,1,1]$-\texttt{BN}-\texttt{ReLU}-\texttt{convres}$[16,3,1,1]$-\texttt{BN}-\texttt{ReLU}-\texttt{convres}$[16,3,1,1]$-\texttt{BN}-\texttt{ReLU}-\texttt{convres}$[32,3,1,1]$-\texttt{BN}-\texttt{ReLU}-\texttt{convres}$[32,3,1,1]$-\texttt{BN}-\texttt{ReLU}-\texttt{convres}$[64,3,1,1]$-\texttt{BN}-\texttt{ReLU}-\texttt{convres}$[64,3,1,1]$-\texttt{BN}-\texttt{ReLU}, and the output of this convolutional layer (size $64$) is connected to a 6-layered feed-forwarding neural network. Each layer in the feed-forwarding network contains the same number of nodes of $200$, and each node in the feed-forwarding network is batch-normalized. 

\end{document}